\documentclass[letterpaper]{article} 
\usepackage{aaai2026}  
\usepackage{times}  
\usepackage{helvet}  
\usepackage{courier}  
\usepackage[hyphens]{url}  
\usepackage{graphicx} 
\usepackage{natbib}  
\usepackage{caption} 
\usepackage{bm}
\usepackage{xcolor}
\usepackage{diagbox}
\usepackage{algorithm}
\usepackage{algorithmic}
\usepackage{amsmath}
\usepackage{booktabs}
\usepackage{xcolor}
\usepackage{multirow}
\usepackage{newfloat}
\usepackage{listings}
\usepackage{float}
\DeclareCaptionStyle{ruled}{labelfont=normalfont,labelsep=colon,strut=off} 
\floatstyle{ruled}
\newfloat{listing}{tb}{lst}{}
\floatname{listing}{Listing}
\title{Adaptive Dynamic Dehazing via Instruction-Driven and Task-Feedback Closed-Loop Optimization for Diverse Downstream Task Adaptation}
\author{
    Yafei Zhang\textsuperscript{\rm 1},
	Shuaitian Song\textsuperscript{\rm 1},
	Huafeng Li\textsuperscript{\rm 1}\equalcontrib,
	Shujuan Wang\textsuperscript{\rm 1},
	Yu Liu\textsuperscript{\rm 2}
}
\affiliations{
    \textsuperscript{\rm 1}Faculty of Information Engineering and Automation, Kunming University of Science and Technology, Kunming 650500, China \\
    \textsuperscript{\rm 2}School of Instrument Seicence and Opto-electronics Engineering, Hefei University of Technology, Hefei 230009 , China \\

 	zyfeimail@163.com, songshuaitiann@163.com, hfchina99@163.com, wangsj@kust.edu.cn, yuliu@hfut.edu.cn
}

\usepackage{bibentry}

\begin{document}

\maketitle

\begin{abstract}
In real-world vision systems, haze removal is required not only to enhance image visibility but also to meet the specific needs of diverse downstream tasks. To address this challenge, we propose a novel adaptive dynamic dehazing framework that incorporates a closed-loop optimization mechanism. It enables feedback-driven refinement based on downstream task performance and user instruction–guided adjustment during inference, allowing the model to satisfy the specific requirements of multiple downstream tasks without retraining. Technically, our framework integrates two complementary and innovative mechanisms: (1) a task feedback loop that dynamically modulates dehazing outputs based on performance across multiple downstream tasks, and (2) a text instruction interface that allows users to specify high-level task preferences. This dual-guidance strategy enables the model to adapt its dehazing behavior after training, tailoring outputs in real time to the evolving needs of multiple tasks. Extensive experiments across various vision tasks demonstrate the strong effectiveness, robustness, and generalizability of our approach. These results establish a new paradigm for interactive, task-adaptive dehazing that actively collaborates with downstream applications.
\end{abstract}

\begin{links}
     \link{Code}{https://github.com/songshuaitian/ADeT-Net}
\end{links}

\section{Introduction}
Haze is a common atmospheric condition that severely impairs image visibility and scene understanding. Consequently, image dehazing has become a critical preprocessing step in real-world systems like autonomous driving and surveillance. Early approaches mainly aimed to enhance visual quality using handcrafted priors or supervised models trained on synthetic data. However, improving visual appearance alone does not guarantee better performance in downstream vision tasks. In practical scenarios, the dehazed output often serves as the input to task-specific models, and misalignment between the objectives of dehazing and those of downstream tasks can lead to suboptimal or even detrimental outcomes. To address this issue, recent studies \cite{1,2} have explored integrating downstream tasks into the dehazing pipeline by jointly training the dehazing model with a specific task network. While such integration can enhance performance for a single task, it also introduces significant limitations: these methods require retraining for each new task and lack flexibility to adapt to different tasks once deployed. Despite these efforts, achieving a generalizable solution that supports diverse downstream tasks remains an open challenge.
\begin{figure}[t!]
	\begin{center}
		\includegraphics[width=0.850\linewidth]{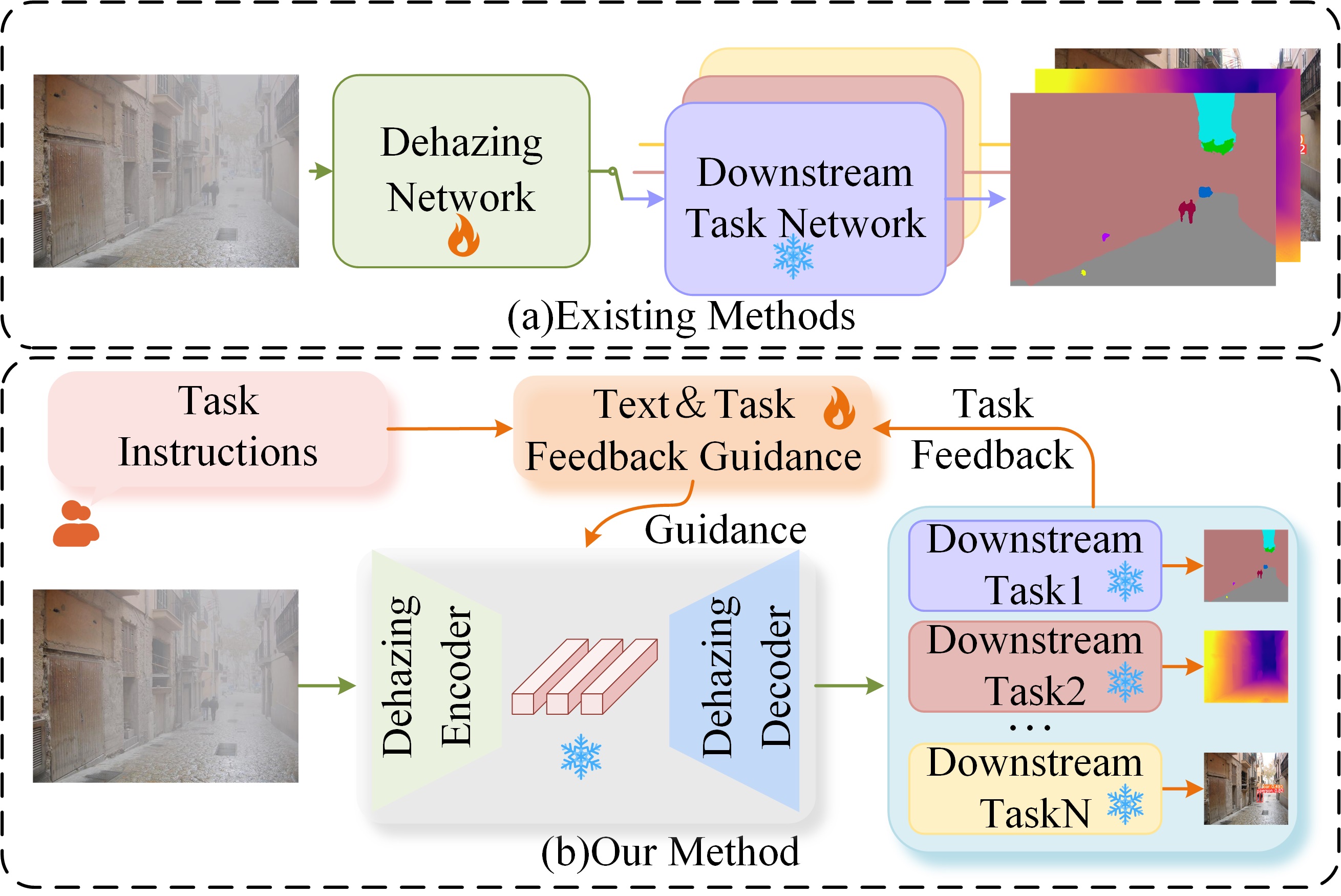}
	\end{center}
	\caption{Comparison between existing methods and ours.}
	\label{fig1}
\end{figure}

To address these limitations, we propose a closed-loop optimization framework for adaptive and dynamic image dehazing. Unlike traditional approaches that produce static outputs regardless of task context (As shown in Figure \ref{fig1}), our method is designed to optimize dehazing results not only in terms of visual clarity but also with respect to the specific requirements of downstream tasks. The key innovation lies in a dual-guidance mechanism that integrates both feedback from downstream task performance and high-level user intent expressed in textual instructions. By jointly leveraging these two sources of guidance, the model can adjust its dehazing behavior in real time during inference, without requiring any retraining or task-specific fine-tuning. This architecture holds promise for serving as a flexible and interactive dehazing module adaptable to various downstream tasks.

To realize the above design, we construct an adaptive dehazing framework centered on a closed-loop optimization mechanism. The process begins with training an Initial Dehazing Network (IDN) on synthetic hazy data to acquire general image restoration capabilities. During inference, the dehazing results are refined in real time through a dual-guidance strategy. Specifically, feedback from downstream task performance is used to guide feature modulation based on how well the current dehazed image supports the task. Meanwhile, user-provided textual instructions are interpreted to capture high-level intent and semantic preferences. These two signals—task-driven feedback and instruction-based semantics—jointly inform the modulation of intermediate features within the dehazing network, enabling dynamic and context-aware adaptation without requiring retraining. This framework introduces a novel paradigm that bridges low-level image restoration with high-level task guidance, allowing the dehazing process to become interactive, controllable, and task-specific. This design offers a generalizable solution for real-world vision systems where adaptability and compatibility with multiple tasks are essential. In summary, the main contributions of this work are as follows:
\begin{itemize}
	\item We propose a novel closed-loop dehazing framework that enables dynamic, task-aware, and instruction-driven refinement during inference, achieving real-time adaptation without any model retraining or fine-tuning. This design significantly improves the flexibility and deployment efficiency of dehazing model in dynamic, multi-task environments.
	
    \item We present a dual-guidance mechanism that combines downstream task feedback and semantic-level textual instructions to support dynamic, task-specific adaptation of the dehazing process. This mechanism is instantiated through two  modules—Task Feedback-Guided Adaptation (TFGA) and Instruction-Guided Modulation (IGM)—which collaboratively enable real-time and fine-grained refinement of dehazing outputs according to varied task objectives, without requiring retraining.

	\item We conduct extensive experiments on diverse downstream tasks including detection, segmentation, and depth estimation. Results show consistent performance gains over both traditional and task-aware baselines, demonstrating the effectiveness and adaptability of our approach, and offering a promising paradigm for interactive and task-driven dehazing in real-world applicaiton.
\end{itemize}

\section{Related Work}
\subsection{Typical Image Dehazing}
In image dehazing, atmospheric scattering model-based methods are widely used. These approaches typically estimate the transmission map and atmospheric light to reconstruct clear images, and can be categorized into three types: end-to-end joint parameter estimation (E2E-JPE) methods \cite{5,6,7}, physical model simplification (PMS) methods \cite{10,11}, and physical prior-guided (PPG) methods \cite{12,13}. E2E-JPE methods, such as DehazeNet \cite{5} and DCPDN \cite{6}, treat transmission and atmospheric light as learnable parameters and substitute them into the scattering model for restoration. They provide high physical interpretability but are sensitive to parameter errors. PMS methods, like AOD-Net \cite{10} and FAMED-Net \cite{11}, simplify the model by combining parameters, but at the cost of physical fidelity. PPG methods utilize priors (e.g., dark/bright channel \cite{12,13}) to guide estimation, offering generalization with limited data but prone to artifacts when assumptions fail.

Beyond model-based approaches, end-to-end deep learning methods directly map hazy images to clear outputs. They are typically CNN-based \cite{15,16,40}, GAN-based \cite{21,22}, or Transformer-based \cite{23,24}. CNN-based methods exploit local features without modeling physical processes, while GAN-based approaches enhance realism via adversarial training but may suffer from instability. Transformer-based methods capture long-range dependencies through self-attention, effectively modeling fog patterns. However, most of these methods overlook the adaptability of dehazed results to downstream tasks.

\begin{figure*}
	\begin{center}
		\includegraphics[width=0.95\linewidth]{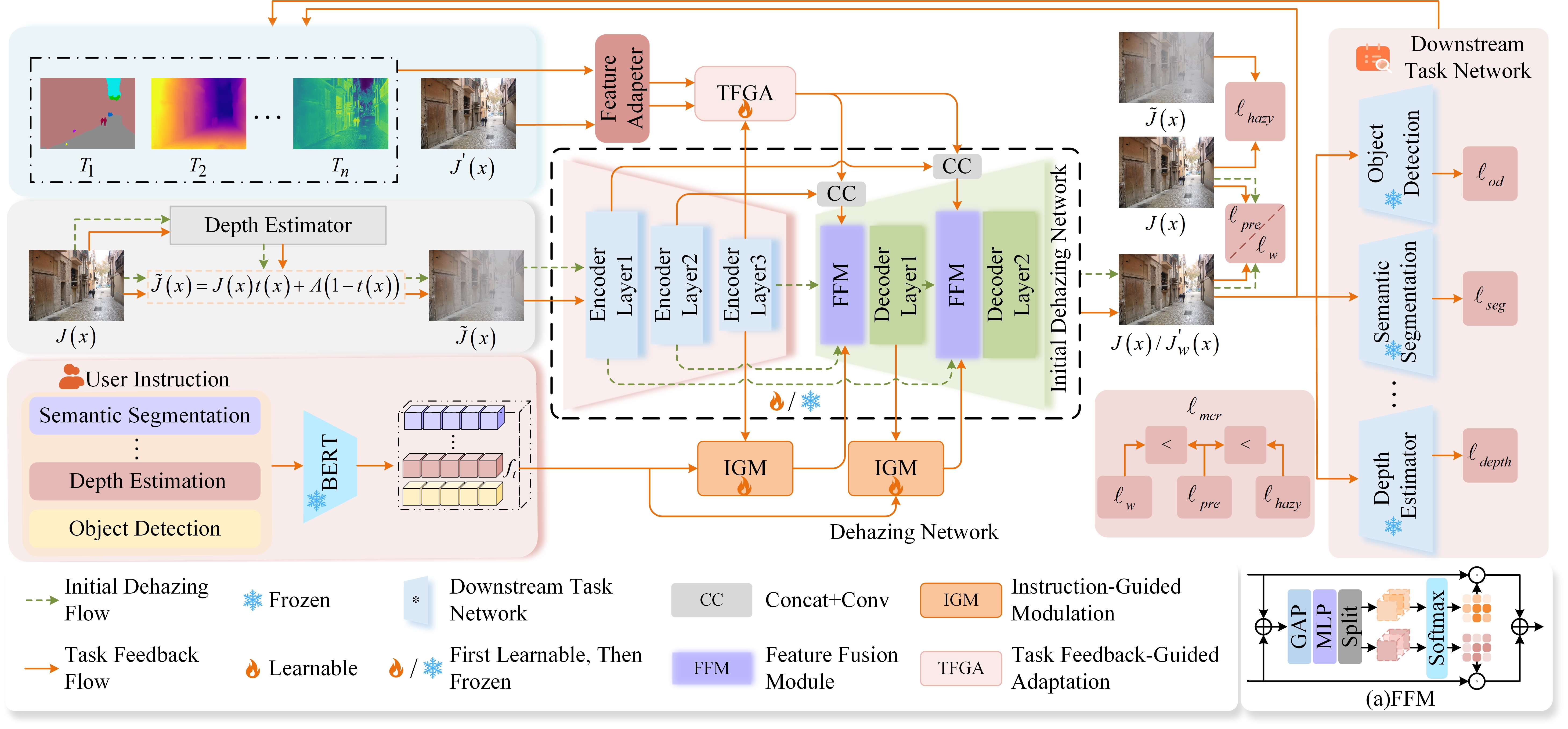}
	\end{center}
	\caption{Overview of the proposed method. The method constructs a closed-loop regulation mechanism jointly guided by semantic task instructions and task performance feedback. It leverages the semantic features of text instructions extracted by BERT, the initial dehazed images, and feedback from downstream tasks to collaboratively adjust dehazing features via the IGM and TFGA modules, enabling adaptive optimization across diverse downstream scenario.}
	\label{fig2}
\end{figure*}

\subsection{Downstream Task-Driven Image Dehazing}
Downstream task-driven image dehazing has garnered increasing attention, yet conventional methods often focus solely on enhancing visual quality while overlooking their impact on downstream tasks. To address this issue, researchers have explored joint optimization strategies that couple dehazing with high-level vision tasks. Methods such as UDnD \cite{1}, ADAM-Dehaze \cite{2}, and MS-FODN \cite{4} integrate object detection into the dehazing process, while VRD-IR \cite{3} incorporates object recognition. Although these approaches consider the influence of dehazing on downstream tasks, they are typically tailored to a single task and lack the flexibility to generalize across multiple tasks. Furthermore, they do not utilize performance feedback or semantic guidance from downstream tasks to regulate the dehazing process, often leading to suboptimal outputs.

\section{Methodology}
\subsection{Overview}
Figure \ref{fig2} illustrates the framework of the proposed method. To adapt the dehazing results to various downstream tasks without requiring task-specific fine-tuning, a closed-loop optimization strategy is introduced, combining text instruction guidance and feedback from downstream tasks. In this framework, the initial dehazed images are input into downstream task models aligned with the objectives described in the text instructions. Feedback from these tasks, together with semantic information extracted from the instructions, jointly informs adjustments to the dehazing network. This enables the network to generate results better suited to the current task's requirements. Two key modules support this dynamic adjustment: the TFGA module, which adapts the network’s feature outputs based on task performance, and the IGM module, which interprets the instructions' semantic content to guide dehazing adjustments. Together, these modules form a closed-loop system that connects dehazing outputs and task execution feedback, thereby enhancing the adaptability of the dehazing results across diverse tasks.

\subsection{Initial Dehazing Netowrk}
As shown in Figure~\ref{fig2}, based on the atmospheric scattering model, we add haze to the clear image $J(x)$ with the assistance of a depth estimator to generate a hazy image $\tilde{J}(x)$. $\tilde{J}(x)$ is then fed into the IDN encoder to extract deep features. These features are then passed to the decoder for image reconstruction, yielding an initial dehazed result $J'(x)$. The IDN, based on Transformer architecture and following the U-Net encoder-decoder paradigm, includes three encoder layers for processing features at different scales. The decoder consists of two layers and two FFMs. Features at the same scale from the encoder and decoder are combined via residual connections and then passed to the FFM for fusion.

During IDN training, the FFM integrates features from the encoder and decoder at corresponding scales, reducing information loss during transmission. In the closed-loop optimization stage, the FFM further receives task semantics from the TFGA and IGM modules. This enables the model to dynamically regulate the dehazing process according to task requirements and instruction content. To ensure strong restoration performance, we jointly optimize the IDN using both $l_1$-loss and contrastive loss:
\begin{equation}
	\footnotesize
	\begin{aligned}
		\ell_{predeh} = &\; \| J'(x) - J(x) \|_1 \\
		& + \lambda \sum_{v=1}^{n} \beta_v \cdot 
		\frac{ \| \text{VGG}_v(J(x)) - \text{VGG}_v(J'(x)) \|_1 }
		{ \| \text{VGG}_v(J'(x)) - \text{VGG}_v(\tilde{J}(x)) \|_1 }
	\end{aligned}
\end{equation}
where $\text{VGG}_v$ denotes the output from the $v$-th layer of the VGG19~\cite{25}, and $\beta_v$ is the corresponding weighting coefficient, as defined in~\cite{26}. In addition, $\lambda$ is a hyperparameter set to 0.1.

\subsection{Task Feedback-Guided Adaptation}

\begin{figure} [t!]
	\begin{center}
		\includegraphics[width=0.90\linewidth]{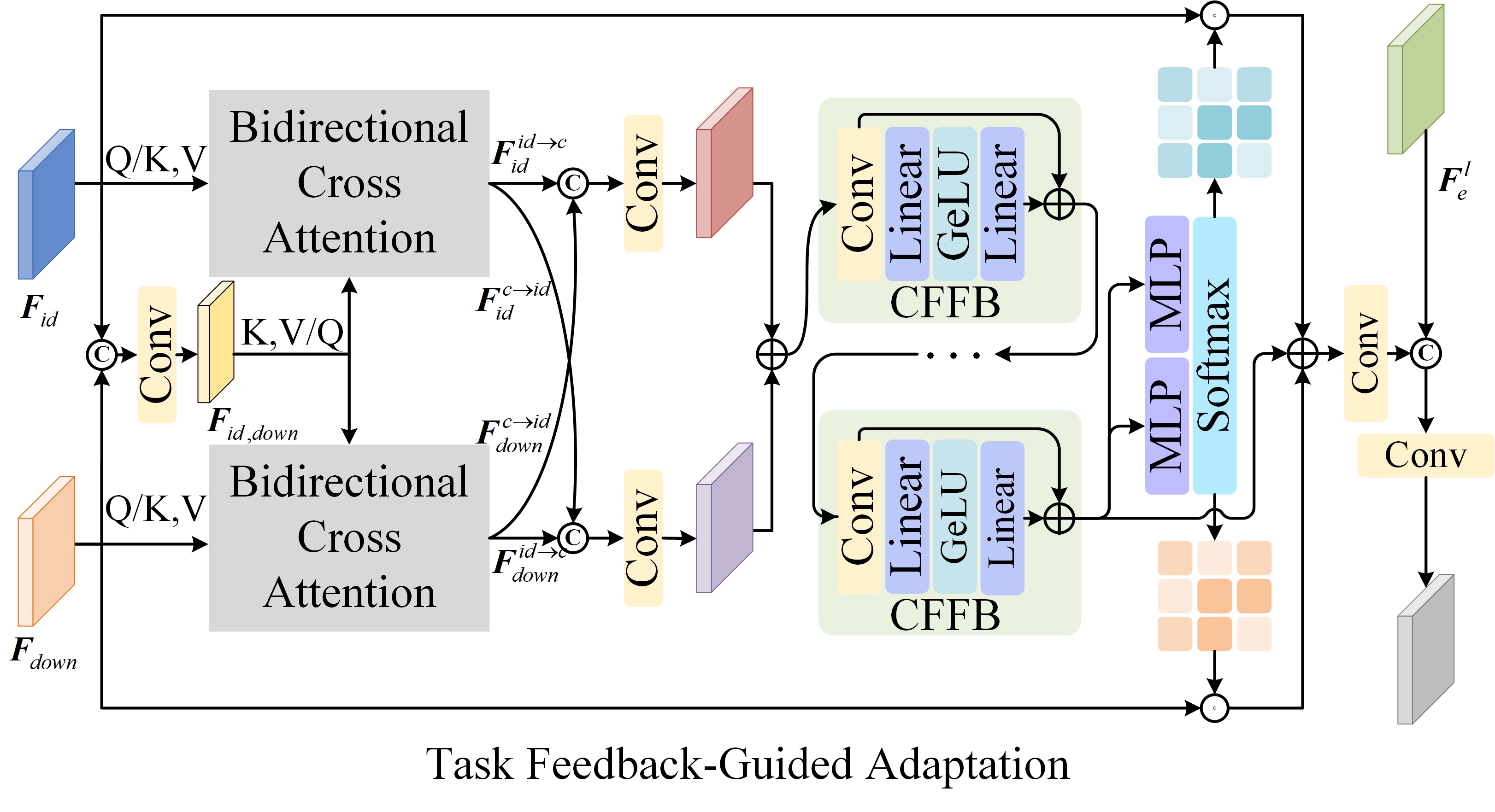}
	\end{center}
	\caption{Structure of the TFGA.}
	\label{fig3}
\end{figure}

Given that the decoder in a dehazing network primarily performs image reconstruction and detail generation, regulating it enables precise optimization of image quality with relatively low complexity and risk due to its relatively independent structure. Therefore, this paper focuses the regulation on the decoder to directly guide detail recovery during image reconstruction, thereby producing dehazed images better aligned with the requirements of downstream tasks. As illustrated in Figure \ref{fig3}, the TFGA is mainly composed of a bidirectional cross-attention mechanism and two Channel-wise Feature Fusion Blocks (CFFB). The input features, $\bm{F}_{id}$ and  $\bm{F}_{down}$, are the outputs of the Feature Adapter applied to $J^{\prime}(x)$ and downstream task feedback, respetively. For tasks such as semantic segmentation or depth estimation, $\bm{F}_{down}$ is the output of the downstream task applied to the dehazed image $J^{\prime}(x)$. For object detection, $\bm{F}_{down}$ refers to the intermediate features extracted by the feature extraction network of the downstream task from $J^{\prime}(x)$.

To enhance the representational capacity of useful information in $\bm{F}_{id}$, the bidirectional cross-attention mechanism performs interaction modeling in two directions. In the first direction, $\bm{F}_{id}$ and $\bm{F}_{down}$ are concatenated and passed through a convolutional layer, followed by a linear transformation to obtain the fused feature $\bm{F}_{id,down}$, which serves as the query $\bm{Q}$ for the attention module. Meanwhile, $\bm{F}_{id}$ is linearly projected to generate the key-value pairs $\bm{K}$ and $\bm{V}$. The resulting cross-attention output is denoted as $\bm{F}_{id}^{id \to c}$. In the reverse direction, $\bm{F}_{id,down}$ is linearly projected to form $\bm{K}$ and $\bm{V}$, which are then used to attend to $\bm{F}_{id}$, producing the output $\bm{F}_{id}^{c \to id}$.  Symmetrically, $\bm{F}_{down}$ undergoes the same procedure to yield $\bm{F}_{down}^{id \to c}$ and $\bm{F}_{down}^{c \to id}$.

To further explore the structural information in $\bm{F}_{id}$ and $\bm{F}_{down}$, we concatenate $\bm{F}_{id}^{id \to c}$ with $\bm{F}_{down}^{c \to id}$, and $\bm{F}_{id}^{c \to id}$ with $\bm{F}_{down}^{id \to c}$. These concatenated features are then processed through convolutional layers to extract deeper semantic representations. The outputs are summed and passed through two CFFBs to obtian $\bm{F}_{idd}$, which is then fed into two Multi-Layer Perceptrons (MLPs). Finally, a Softmax function is applied to obtain the regulation weight matrices $\bm{Q}_{id}$ and $\bm{Q}_{down}$ for $\bm{F}_{id}$ and $\bm{F}_{down}$. The fused feature incorporating downstream task feedback is computed as:
\begin{equation}
	\footnotesize
	\bm{F}_{id,dow} = \mathrm{Conv}\left(\bm{F}_{id} \odot \bm{Q}_{id} + \bm{F}_{down} \odot \bm{Q}_{down} + \bm{F}_{idd}\right)
	\label{eq:feedback_fusion}
\end{equation}
We concatenate $\bm{F}_{id,dow}$ with the output $\bm{F}_e^l$ from the final Transformer layer in the encoder, and further process them with a convolutional layer to obtain the modulated result that reflects the downstream task feedback. This result, along with $\bm{F}_e^l$ and the output of the previous encoder layer, is input into the FFM for integration. The integrated feature is then used in the decoding stage.

In the aforementioned process, the performance of the downstream task network on the initial dehazed result $J^{\prime}(x)$ is fed back to the dehazing network. This feedback then guides the feature adjustment during the dehazing process, enabling the representations extracted at the decoder end to be more aligned with the requirements of specific tasks. Through this feedback mechanism, the dehazing network not only possesses the capability to restore clear images but also generates feature representations that are more conducive to downstream task.

\begin{figure}
	\begin{center}
		\includegraphics[width=0.90\linewidth]{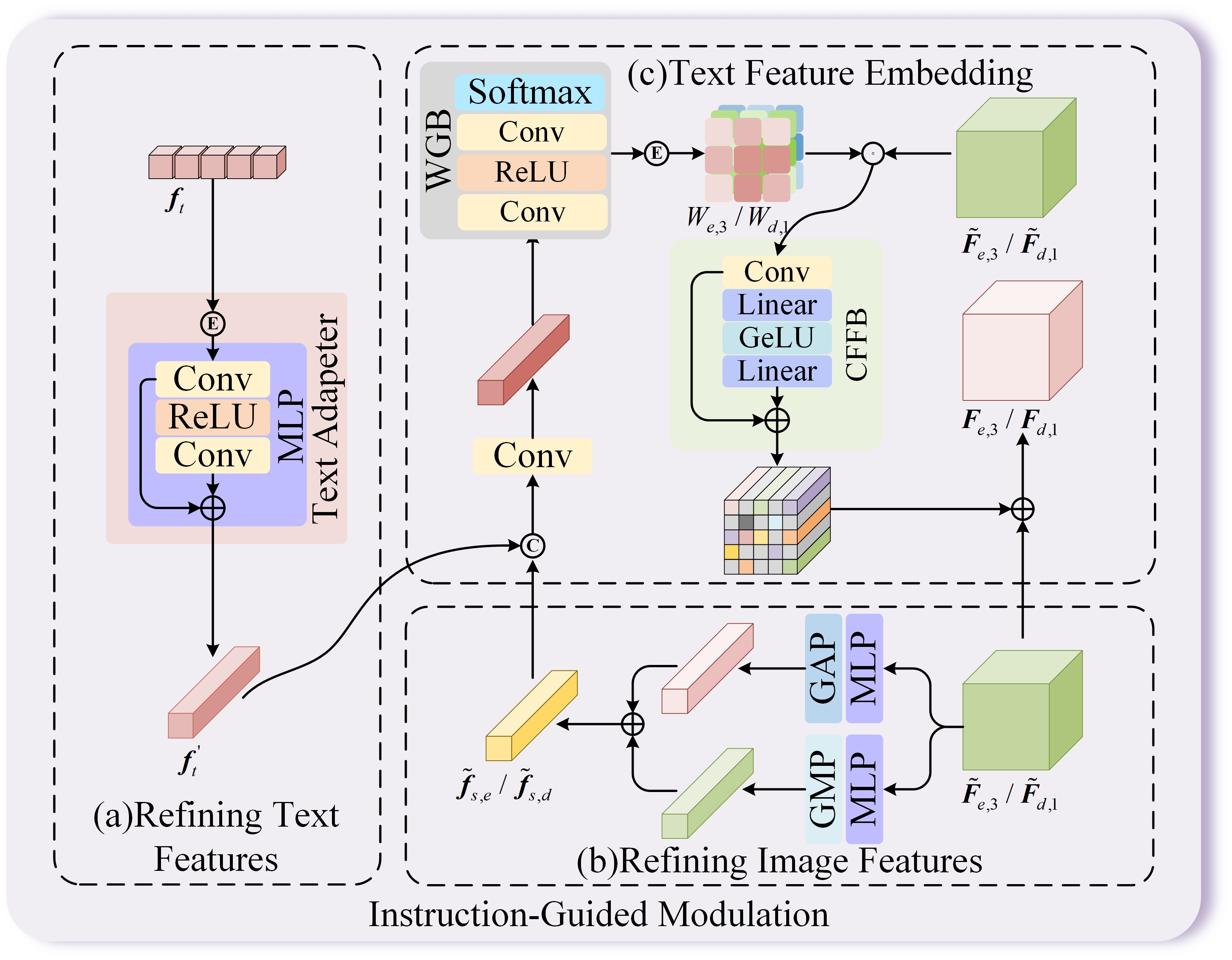}
	\end{center}
	\caption{Structure of the IGM.}
	\label{fig4}
\end{figure}

\subsection{Instruction-Guided Modulation}
Although downstream task feedback can make the restoration results more task-specific, the optimization process may become blind and inefficient without semantic guidance. Therefore, this paper further introduces textual instruction information to incorporate semantic guidance alongside task feedback, establishing a complementary relationship between semantic intention parsing and performance-driven regulation.  Specifically, the IGM interprets the task semantics from the text, while the TFGA handles feedback from downstream tasks. This approach forms a robust closed-loop optimization mechanism, improving the adaptability of dehazed images to diverse downstream tasks.

To more directly leverage the textual instructions for enhancing the controllability and task adaptability of the model, the output of the IGM is applied exclusively to the decoder of the dehazing network. Specifically, the task instructions (\texttt{text}) provided by the user are fed into a pre-trained BERT model to extract the instruction feature vector $\bm{f}_t$. Let $\bm{\tilde{F}}_{e,3}$ denote the features obtained from the third layer of the encoder in the IDN, and let $\bm{\tilde{F}}_{d,1}$ represent the features reconstructed by the first layer of the decoder. In our method, two IGM modules are introduced to gradually inject task semantics into the decoding process, thereby equipping the dehazing network with a degree of semantic controllability. The first IGM takes $\bm{f}_t$ and $\bm{\tilde{F}}_{e,3}$ as inputs to guide the adjustment of decoder features, while the second IGM uses $\bm{f}_t$ and $\bm{\tilde{F}}_{d,1}$ to further enhance the embedding of task-specific information. As illustrated in Figure~\ref{fig4}, the IGM module comprises three key components: text feature refinement, image feature refinement, and text feature embedding.

\begin{figure*}[t!]
	\begin{center}
		\includegraphics[width=0.93\linewidth]{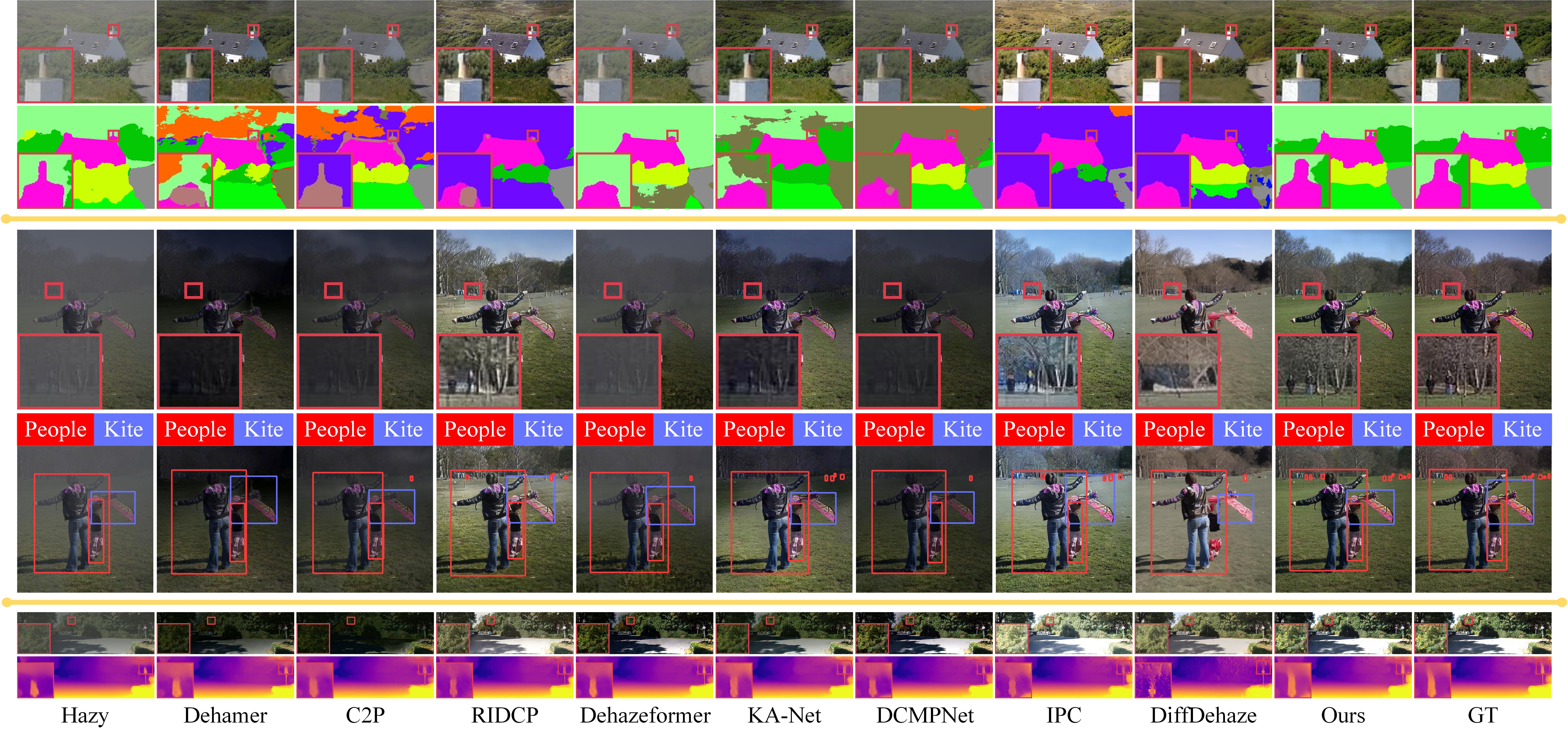}
	\end{center}
	\caption{Visual comparison with state-of-the-art methods on Setting 1. Each part includes two rows: dehazing results (rows 1) and corresponding downstream task outputs (rows 2). Input images are taken from ADE20K, COCO, and KITTI.}
	\label{fig5}
\end{figure*}

The core objective of text feature refinement is to refine a task-oriented representation from the BERT-extracted feature $\bm{f}_t$. This process is implemented using a text adapter composed of an expand operation and a MLP. The adapter not only enhances the representation of task-related semantics but also projects $\bm{f}_t$ from the text semantic space into the image feature space, facilitating cross-modal alignment and feature modulation. The output of the adapter is denoted as $\bm{f}'_t$. The image feature refinement block comprises two branches, each consisting of an MLP followed by either GMP or GAP, respectively. The outputs of these two branches are summed to obtain the refined image feature representation. When the inputs to the IGM are $\bm{f}_t$ and $\bm{\tilde{F}}_{e,3}$ or $\bm{\tilde{F}}_{d,1}$, the corresponding refined results are denoted as $\bm{\tilde{f}}_{s,e}$ and $\bm{\tilde{f}}_{s,d}$, respectively.

The text feature embedding block consists of a Weight Generation Block (WGB) and a CFFB, as shown in Figure~\ref{fig4}. During embedding, $\bm{f}'_t$ and $\bm{\tilde{f}}_{s,e}$ / $\bm{\tilde{f}}_{s,d}$ are input into the WGB, which generates modulation parameters $\bm{W}_{e,3}$ ($\bm{W}_{d,1}$) through an expand operation. These parameters modulate the input image features $\bm{\tilde{F}}_{e,3}$ ($\bm{\tilde{F}}_{d,1}$):
\begin{equation}
	\footnotesize
	\bm{\bar{F}}_{e,3} = \bm{W}_{e,3}  \odot \bm{\tilde{F}}_{e,3} ,  \quad \bm{\bar{F}}_{d,1} = \bm{W}_{d,1}  \odot \bm{\tilde{F}}_{d,1} 
\end{equation}

The modulated features $\bm{\bar{F}}_{e,3}$ ($\bm{\bar{F}}_{d,1}$) are subsequently fed into the CFFB. The output is added to the original image features to obtain the final modulated representations:
\begin{equation}
	\footnotesize
	\bm{F}_{e,3} = \text{CFFB}(\bm{\bar{F}}_{e,3}) + \bm{\tilde{F}}_{e,3} , \quad \bm{F}_{d,1} = \text{CFFB}(\bm{\bar{F}}_{d,1}) + \bm{\tilde{F}}_{d,1}
\end{equation}
In this process, the CFFB further enhances the structural expressiveness of the visual features $\bm{\tilde{F}}_{e,3}$ ($\bm{\tilde{F}}_{d,1}$), thereby highlighting information beneficial for image restoration. Furthermore, since the semantic information from the textual instructions is injected into $\bm{F}_{e,3}$ and $\bm{F}_{d,1}$, the visual features can be dynamically adapted to downstream task instructions.

To ensure that the dehazed results modulated by the TFGA and the IGM better align with the requirements of downstream tasks, we use both $l_1$-loss and contrastive loss:
\begin{equation}
	\footnotesize
	\begin{array}{l}
		\ell_{dehaze} = \|J'_w(x) - J(x)\|_1 + \\
		\quad \lambda \sum_{v=1}^{n} \beta_v \frac{\|VGG_v(J(x)) - VGG_v(J'_w(x))\|_1}{\|VGG_v(J'_w(x)) - VGG_v(\tilde{J}(x))\|_1}
	\end{array}
\end{equation}
where $J'_w(x)$ denotes the dehazed result after modulation. We expect this modulated result, guided by downstream task feedback and text instructions, to outperform the intermediate and initial dehazed outputs. Accordingly, the reconstruction loss is expected to satisfy the following inequality:
\begin{equation}
	\footnotesize
	\|J'_w(x) - J(x)\|_1 < \|J'(x) - J(x)\|_1 < \|\tilde{J}(x) - J(x)\|_1
\end{equation}
To enforce the relative quality constraint between different dehazing results, we propose a Multi-level Contrastive Ranking Loss ($\ell_{mcr}$), defined as:
\begin{equation}
	\footnotesize
	\ell_{mcr} = \max(\ell_w - \ell_{p} + \beta_1, 0) + \max(\ell_w - \ell_{h} + \beta_2, 0)
\end{equation}
where $\beta_1$ and $\beta_2$ are hyperparameters, empirically set to 0.1 and 0.3, respectively, with $\beta_1 < \beta_2$. The individual loss terms are defined as follows: $\ell_w = \|J'_w(x) - J(x)\|_1$, 
$\ell_{p} = \|J'(x) - J(x)\|_1$, 
and 
$\ell_{h} = \|\tilde{J}(x) - J(x)\|_1$.

\begin{table*}[ht!]
	\centering\small
	\renewcommand{\arraystretch}{1.1}%
	\resizebox{0.9\textwidth}{!}{%
		\begin{tabular}{l|ccc|ccc|ccc}
			\hline
			\multirow{2}{*}{Methods} & \multicolumn{3}{c|}{ADE20K} & \multicolumn{3}{c|}{COCO} & \multicolumn{3}{c}{KITTI} \\
			\cline{2-10}
			& PSNR$\uparrow$ & SSIM$\uparrow$ & LPIPS$\downarrow$ & PSNR$\uparrow$ & SSIM$\uparrow$ & LPIPS$\downarrow$ & PSNR$\uparrow$ & SSIM$\uparrow$ & LPIPS$\downarrow$ \\
			\hline
			Dehamer \cite{23} & 22.63 & 0.8964 & 0.1538 & 23.26 & 0.9280 & 0.1369 & 24.87 & 0.9212 & 0.1029 \\
			C2P \cite{16} & 22.26 & 0.8842 & 0.1594 & 23.02 & 0.9044 & 0.1322 & 25.01 & 0.9342 & 0.0926 \\
			RIDCP \cite{37} & \underline{27.21} & 0.9207 & 0.0702 & \textbf{28.18} & 0.8990 & 0.0974 & 27.18 & 0.9441 & 0.1242 \\
			Dehazeformer \cite{24} & 23.76 & 0.8958 & 0.1332 & 22.41 & 0.8639 & 0.1412 & 23.36 & 0.9160 & 0.1163 \\
			KA-Net \cite{38} & 24.57 & 0.9153 & 0.0914 & 24.92 & 0.9241 & 0.1291 & 26.73 & 0.9413 & 0.0968 \\
			DCMPNet \cite{17} & 25.64 & \underline{0.9571} & \underline{0.0326} & 25.39 & \underline{0.9422} & 0.0892 & 28.52 & 0.9617 & 0.0384 \\
			DiffDehaze \cite{19} & 25.43 & 0.9194 & 0.0853 & 24.68 & 0.9127 & 0.1048 & 26.91 & 0.9568 & 0.0454 \\
			IPC  \cite{39} & 26.23 & 0.9432 & 0.0618 & 26.32 & 0.9381 & \underline{0.0837} & \underline{28.94} & \underline{0.9632} & \underline{0.0296} \\
			\textbf{Ours} & \textbf{27.47} & \textbf{0.9701} & \textbf{0.0293} & \underline{27.14} &\textbf{0.9587} & \textbf{0.0698} & \textbf{30.50} & \textbf{0.9740} & \textbf{0.0176} \\
			\hline
	\end{tabular}}
	\caption{Comparison of dehazing performance. Best and runner-up values are highlighted in bold and underlined, respectively.} 
	\label{tab:1}
	
\end{table*}

\begin{table*}[ht!]
	\centering
	\renewcommand{\arraystretch}{1.1}
	\resizebox{0.9\textwidth}{!}{%
		\begin{tabular}{l|c|cc|cccc|ccc}
			\hline
			\multirow{2}{*}{Methods} & \multicolumn{1}{c|}{SS} & \multicolumn{2}{c|}{OD} & \multicolumn{4}{c|}{DE (Error Metric $\downarrow$)} & \multicolumn{3}{c}{DE (Accuracy Metric $\uparrow$)} \\
			\cline{2-11}
			& mIoU$\uparrow$ & mAP$\uparrow$ & mAP50-95$\uparrow$ & AbsRel & SqRel & RMSE & RMSElog & $\delta<1.25$ & $\delta<1.25^2$ & $\delta<1.25^3$ \\
			\hline
			Dehamer & 44.04 & 49.8 & 30.4 & 0.115 & 0.860 & 4.802 & 0.195 & 0.874 & 0.956 & 0.980 \\
			C2P & 46.66 & 52.3 & 33.5 & 0.123 & 0.888 & 4.933 & 0.206 & 0.855 & 0.943 & 0.972 \\
			RIDCP & 46.15 & 51.7 & 33.0 & \underline{0.100} & \underline{0.680} & \underline{4.307} & \underline{0.175} & \underline{0.898} & \underline{0.966} & \textbf{0.984} \\
			Dehazeformer & 45.84 & 52.0 & 33.2 & 0.105 & 0.711 & 4.431 & 0.182 & 0.889 & 0.964 & \underline{0.983} \\
			KA-Net & 45.47 & 50.5 & 31.2 & 0.108 & 0.749 & 4.554 & 0.187 & 0.884 & 0.960 & 0.982 \\
			DCMPNet & \underline{46.92} & \underline{53.1} & \underline{34.2} & 0.108 & 0.725 & 4.579 & 0.190 & 0.883 & 0.959 & 0.980 \\
			DiffDehaze & 41.84 & 42.1 & 26.1 & 0.120 & 0.849 & 5.400 & 0.204 & 0.849 & 0.950 & 0.980 \\
			IPC & 45.90 & 52.3 & 33.9 & \textbf{0.099} & 0.688 & \textbf{4.273} & \textbf{0.174} & \textbf{0.902} & \textbf{0.967} & \textbf{0.984} \\
			\textbf{Ours} & \textbf{50.34} & \textbf{54.7} & \textbf{35.7} & \textbf{0.099} & \textbf{0.662} & 4.310 & \textbf{0.174} & \underline{0.898} & \textbf{0.967} & \textbf{0.984} \\
			\hline
	\end{tabular}}
	\caption{Comparison of downstream task performance across semantic segmentation (SS), object detection (OD), and depth estimation (DE) on Setting 1.  Best and runner-up values are highlighted in bold and underlined, respectively.}
	\label{tab:2}

\end{table*} 

This paper aims to adapt the restoration results of a dehazing model to various downstream tasks without retraining it. Thus, we introduce a task-specific loss term $\ell_{down}$ to ensure the performance of the downstream tasks:
\begin{equation}
	\footnotesize
	\ell_{down} = \ell_{task}(\tilde{y}_{gt}(x), \tilde{y}'(x))
\end{equation}
where $\ell_{task}$ denotes the loss function of the downstream task network. $\tilde{y}_{gt}(x)$ represents the ground truth, and $\tilde{y}'(x)$ denotes the output of $J'_w(x)$ appilied to the downstream task network. For semantic segmentation, object detection, and depth estimation, we adopt the same loss functions as those used in SegFormer~\cite{27}, YOLOv5 \footnote{\url{https://github.com/ultralytics/yolov5}}, and RADepth~\cite{28}, respectively. Accordingly, the overall loss function is defined as:
\begin{equation}
	\footnotesize
	\ell_{total} = \ell_{dehaze} + \ell_{mcr} + \gamma \ell_{down}
\end{equation}
where $\gamma$ is a weighting coefficient for the downstream task loss, empirically set to 0.01. 

\begin{figure*}[t]
	\begin{center}
		\includegraphics[width=0.90\linewidth]{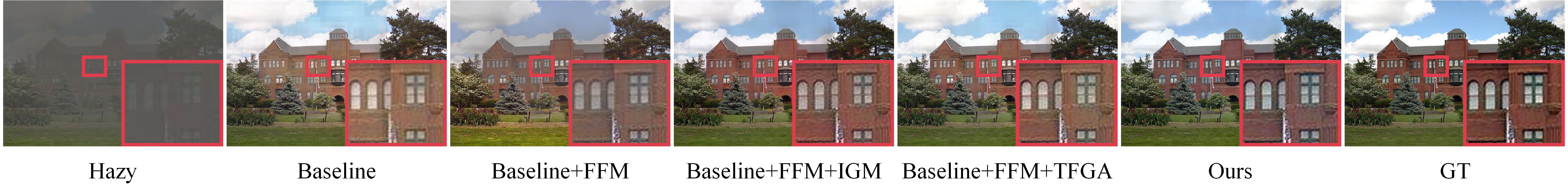}
	\end{center}
	\caption{Visual comparison of dehazing results from ablation studies.}
	\label{fig6}
\end{figure*}
\subsection{Closed-Loop Optimization}

During model inference, the core advantage of our method lies in its ability to dynamically adjust the initial dehazed results through a closed-loop optimization mechanism, thereby achieving better alignment with specific downstream tasks. Unlike traditional dehazing models that operate in a static manner, our approach regulates the dehazing process in real time by leveraging feedback from downstream task performance and user instructions. Such a mechanism not only enhances the capacity of the  model for task-oriented collaboration but also establishes a novel paradigm for building interactive and controllable intelligent visual systems.

\begin{table*}[htbp]
	\centering
	\renewcommand{\arraystretch}{1.1}
	\resizebox{0.9\textwidth}{!}{%
		\begin{tabular}{c|ccc|ccc|ccc}
			\hline
			\multirow{2}{*}{Methods} & \multicolumn{3}{c|}{ADE20K} & \multicolumn{3}{c|}{COCO} & \multicolumn{3}{c}{KITTI} \\
			\cline{2-10}
			& PSNR$\uparrow$ & SSIM$\uparrow$ & LPIPS$\downarrow$ & PSNR$\uparrow$ & SSIM$\uparrow$ & LPIPS$\downarrow$ & PSNR$\uparrow$ & SSIM$\uparrow$ & LPIPS$\downarrow$ \\
			\hline
			Baseline                         & 23.47 & 0.9255 & 0.0791 & 22.83 & 0.9058 & 0.1151 & 25.37 & 0.9091 & 0.0701 \\
			Baseline + FFM                            & 25.27 & 0.9522 & 0.0460 & 24.73 & 0.9368 & 0.0860 & 27.08 & 0.9333 & 0.0353 \\
			Baseline + FFM + IGM                      & 26.36 & 0.9646 & 0.0367 & 25.89 & 0.9515 & 0.0749 & 29.00 & 0.9664 & 0.0241 \\
			Baseline + FFM + TFGA              & 26.62 & 0.9672 & 0.0333 & 26.42 & 0.9550 & 0.0741 & 29.30 & 0.9688 & 0.0223 \\
			Full Model (All Modules)        & \textbf{27.47} & \textbf{0.9701} & \textbf{0.0293} & \textbf{27.14} & \textbf{0.9587} & \textbf{0.0698} & \textbf{30.50} & \textbf{0.9740} & \textbf{0.0176} \\
			\hline  
	\end{tabular}}
	\caption{Ablation studies on the dehazing performance.}
	\label{tab:3}
\end{table*}  

\begin{table*}[ht!]
	\centering
	\renewcommand{\arraystretch}{1.1}
	\resizebox{0.9\textwidth}{!}{%
		\begin{tabular}{c|c|cc|cccc|ccc}
			\hline
			\multirow{2}{*}{Methods} & \multicolumn{1}{c|}{SS} & \multicolumn{2}{c|}{OD} & \multicolumn{4}{c|}{DE (Error Metric $\downarrow$)} & \multicolumn{3}{c}{DE (Accuracy Metric $\uparrow$)} \\
			\cline{2-11}
			& mIoU$\uparrow$ & mAP$\uparrow$ & mAP50-95$\uparrow$ & AbsRel & SqRel & RMSE & RMSElog & $\delta{<}1.25$ & $\delta{<}1.25^2$ & $\delta{<}1.25^3$ \\
			\hline
			Baseline & 49.42 & 53.5 & 34.8 & 0.113 & 0.753 & 4.539 & 0.172 & 0.874 & 0.941 & 0.982 \\
			Baseline + FFM & 50.25 & 54.6 & 35.6 & 0.100 & 0.666 & 4.321 & 0.175 & 0.897 & 0.966 & \textbf{0.984} \\
			Baseline + FFM + IGM & 50.28 & 54.6 & 35.5 & 0.100 & 0.664 & 4.318 & 0.175 & \textbf{0.898} & 0.966 & \textbf{0.984} \\
			Baseline + FFM + TFGA & 50.29 & 54.6 & \textbf{35.7} & 0.100 & 0.663 & 4.316 & 0.175 & \textbf{0.898} & \textbf{0.967} & \textbf{0.984} \\
			Full Model (All Modules) & \textbf{50.34} & \textbf{54.7} & \textbf{35.7} & \textbf{0.099} & \textbf{0.662} & \textbf{4.310} & \textbf{0.174} & \textbf{0.898} & \textbf{0.967} & \textbf{0.984} \\
			\hline
	\end{tabular}}
	\caption{Ablation experiments on downstream tasks.}
	\label{tab:4}
	
\end{table*}

\section{Experiments}
\subsection{Experimental Settings}
\textbf{Datasets}. We train our model using 5,000 randomly sampled images from each of the ADE20K\cite{29}, COCO\cite{30}, and KITTI\cite{31} datasets, where hazy-clear image pairs are synthesized based on the atmospheric scattering model. For evaluation, we use the original test sets, which consist of 2,000, 5,000, and 697 images, respectively.

\textbf{Implementation Details}. We first train the IDN and freeze its parameters upon convergence. Then, only the TFGA and IGM modules are trained. Both stages adopt the Adam optimizer \cite{32} with $\beta_1 = 0.9$, $\beta_2 = 0.999$, an initial learning rate of $1.0 \times 10^{-4}$, and a cosine annealing schedule. To enhance sample diversity, we apply data augmentation strategies similar to those used in SegFormer, YOLOv5, and RADepth across all three training sets. The initial dehazing stage is trained for 300 epochs, followed by 100 epochs for the task feedback stage. All experiments are conducted using PyTorch 1.8.0 on an NVIDIA GeForce RTX 3090 GPU with 24 GB of memory.

\textbf{Evaluation Metrics}. Following existing methods, we adopt PSNR, SSIM, and LPIPS \cite{33} to evaluate dehazing quality. To further demonstrate the effectiveness of our method for downstream tasks, we use the following metrics: Mean Intersection-over-Union (mIoU) \cite{34} for semantic segmentation, Mean Average Precision (mAP) and mAP@[\text{0.5}:0.95) \cite{35} for object detection, and four error metrics (AbsRel, SqRel, RMSE, and RMSElog) and along with three accuracy metrics ($\delta < 1.25$, $\delta < 1.25^2$, and $\delta < 1.25^3$) \cite{36} for depth estimation.

\subsection{Comparison with State-of-the-arts}
We compare the proposed method with eight state-of-the-art approaches: Dehamer, C2P, RIDCP, Dehazeformer, KA-Net, DCMPNet, DiffDehaze, and  IPC.  Specifically, we design three experimental settings: \textbf{Setting 1} trains specific downstream tasks using the dehazing results from each comparison method; \textbf{Setting 2} directly inputs the dehazed results into downstream task networks for testing; \textbf{Setting 3} fine-tunes the dehazing network based on the performance feedback from the downstream tasks.  The results for Setting 1 are shown in Figure~\ref{fig5}, Table~\ref{tab:1}, and Table\ref{tab:2}, while those for \textbf{Settings 2} and  \textbf{3} are in the \textbf{supplementary materials}.


As shown in Figure~\ref{fig5}, our method demonstrates clear advantages in detail preservation, color fidelity, and brightness restoration. The dehazed images produced by our method exhibit visual quality closer to the ground truth (GT). Moreover, the results for downstream tasks in Figure~\ref{fig5} show that our method matches or even surpasses the performance of models specifically trained for each task. This demonstrates its strong adaptability to the
diverse requirements of various downstream tasks. We further conduct quantitative evaluations using objective metrics, as summarized in Table~\ref{tab:1}. Our method achieves the best performance on most metrics, validating its superiority. Quantitative results in Table \ref{tab:2}, obtained by applying downstream task models to the dehazed outputs of different methods, further demonstrate the strong adaptability of our method.

\subsection{Ablation Studies}
The proposed method consists of three modules: TFGA, IGM, and FFM. To evaluate the contribution of each module to overall model performance, we conduct ablation studies using the ADE20K, COCO, and KITTI datasets. Specifically, the baseline model is obtained by removing both TFGA and IGM from the complete model and replacing the FFM module with a simple summation. The ablation experiment results are presented in Figure~\ref{fig6} and Tables \ref{tab:3} and \ref{tab:4}.

\textbf{Effectiveness of FFM}: As shown in Table \ref{tab:3} and \ref{tab:4}, introducing FFM into the baseline model results in the noticeable gains across all evaluation metrics. These results demonstrate the effectiveness of FFM in feature fusion.

\textbf{Effectiveness of IGM}: 
To assess the impact of IGM on model performance, we replace it with a conventional structure consisting of feature concatenation followed by convolution. The results in Table \ref{tab:4} show that incorporating IGM into the baseline+FFM configuration improves performance across all downstream tasks, confirming its effectiveness in enhancing task adaptability.

\textbf{Effectiveness of TFGA}: 
To verify the effectiveness of TFGA, we replace it with a simple addition operation. As indicated in Table \ref{tab:3} and \ref{tab:4}, adding TFGA to the Baseline+FFM setting further improves performance on various metrics, demonstrating its effectiveness in task-guided optimization. In addition, as shown in Figure~\ref{fig6}, the introduction of each module also improves the quality of the dehazed image, further confirming the effectiveness of each module.

\section{Conclusion}
This paper presents a novel adaptive dehazing framework that overcomes the limitations of conventional methods through a dynamic, task-aware, and instruction-driven design. Unlike existing methods that generate static dehazing results, our approach supports real-time adaptation without model retraining, making it well suited for deployment in dynamic, multi-task environments. The framework incorporates a closed-loop optimization strategy supported by two complementary modules: TFGA and IGM. Together, they form a dual-guidance mechanism that enables the model to adaptively tailor its outputs to the specific requirements of diverse downstream tasks during inference. Extensive experiments on object detection, semantic segmentation, and depth estimation demonstrate the effectiveness, robustness, and generalizability of our approach. While our method supports joint modeling and adaptation across multiple downstream tasks, it has only been evaluated on a fixed set of tasks. In real-world scenarios, task types and requirements may change dynamically, posing greater adaptation challenges. Future work will focus on exploring the model's generalization and adaptability to evolving task compositions.

\section{Acknowledgments}
This work was supported in part by the National Science Foundation of China under Grant 62571222 and Grant 62161015, the Yunnan Fundamental Research Projects under Grant 202301AV070004 and Grant 202501AS070123, the Major Science and Technology Special Projects of Yunnan Province under Grant 202502AD080006.

\bibliography{aaai2026}

\clearpage

\twocolumn[
\vspace{2cm}
\begin{center}
	\LARGE\bfseries
	Supplementary Materials for ``Adaptive Dynamic Dehazing via Instruction-Driven and Task-Feedback Closed-Loop Optimization for Diverse Downstream Task Adaptation''
\end{center}
\vspace{2cm}
]
\setcounter{secnumdepth}{0}
\setcounter{section}{0}
\renewcommand{\thesection}{\arabic{section}}

\section{Methodology}
\subsection{1.Closed-Loop Optimization}

During the model inference stage, the core advantage of our method lies in its ability to dynamically adjust the initial dehazed results via a closed-loop optimization mechanism, enabling better alignment with specified downstream tasks, as illustrated in Figure \ref{fig5}. This mechanism overcomes the limitations of traditional dehazing models, which typically operate in a static manner, and empowers our model to regulate the dehazing process in real time based on feedback from downstream tasks and user instructions. Specifically, the TFGA generates modulation signals based on the output of the downstream task model and the initial dehazed results. Meanwhile, the IGM, guided by textual task instructions, provides semantic-level modulation information. These two modules work collaboratively by modifying the input to the FFM module of the initial dehazing network, thereby modulating the intermediate decoder features from two perspectives: task feedback and user instructions. This joint modulation ensures that the final dehazed results are better aligned with the semantic requirements of the current downstream task. A notable advantage of this closed-loop mechanism is its ability to operate independently during the model-frozen testing phase, enabling adaptive feature-level optimization without retraining the IDN or downstream task models.

\begin{figure}[!htpb]
	\begin{center}
		\includegraphics[width=0.850\linewidth]{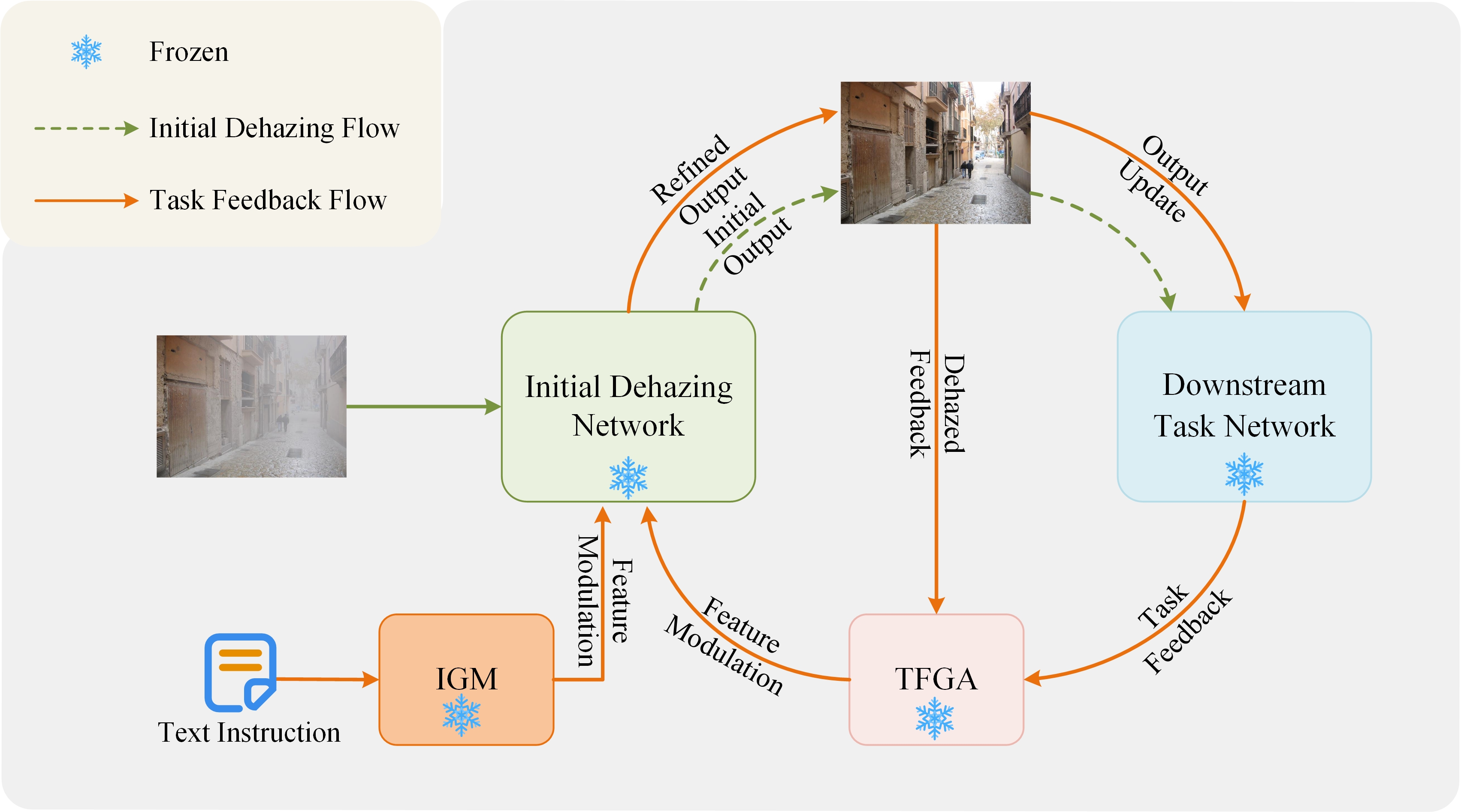}\vspace{-2mm}
	\end{center}
	\caption{Closed-loop optimization framework in the inference phase.}\vspace{-2mm}
	\label{fig11}
\end{figure}

As illustrated in Figure~\ref{fig11}, the closed-loop process comprises four key steps: \textit{Initial Dehazing $\rightarrow$ Task Feedback $\rightarrow$ Feature Modulation $\rightarrow$ Result Update}. This mechanism not only enhances the model’s capability for task collaboration, but also provides a novel paradigm for building interactive and controllable intelligent visual systems.

\section{Experiments}
\subsection{1.Computational Complexity Analysis}
\begin{figure}[!htpb]
	\begin{center}
		\includegraphics[width=0.95\linewidth]{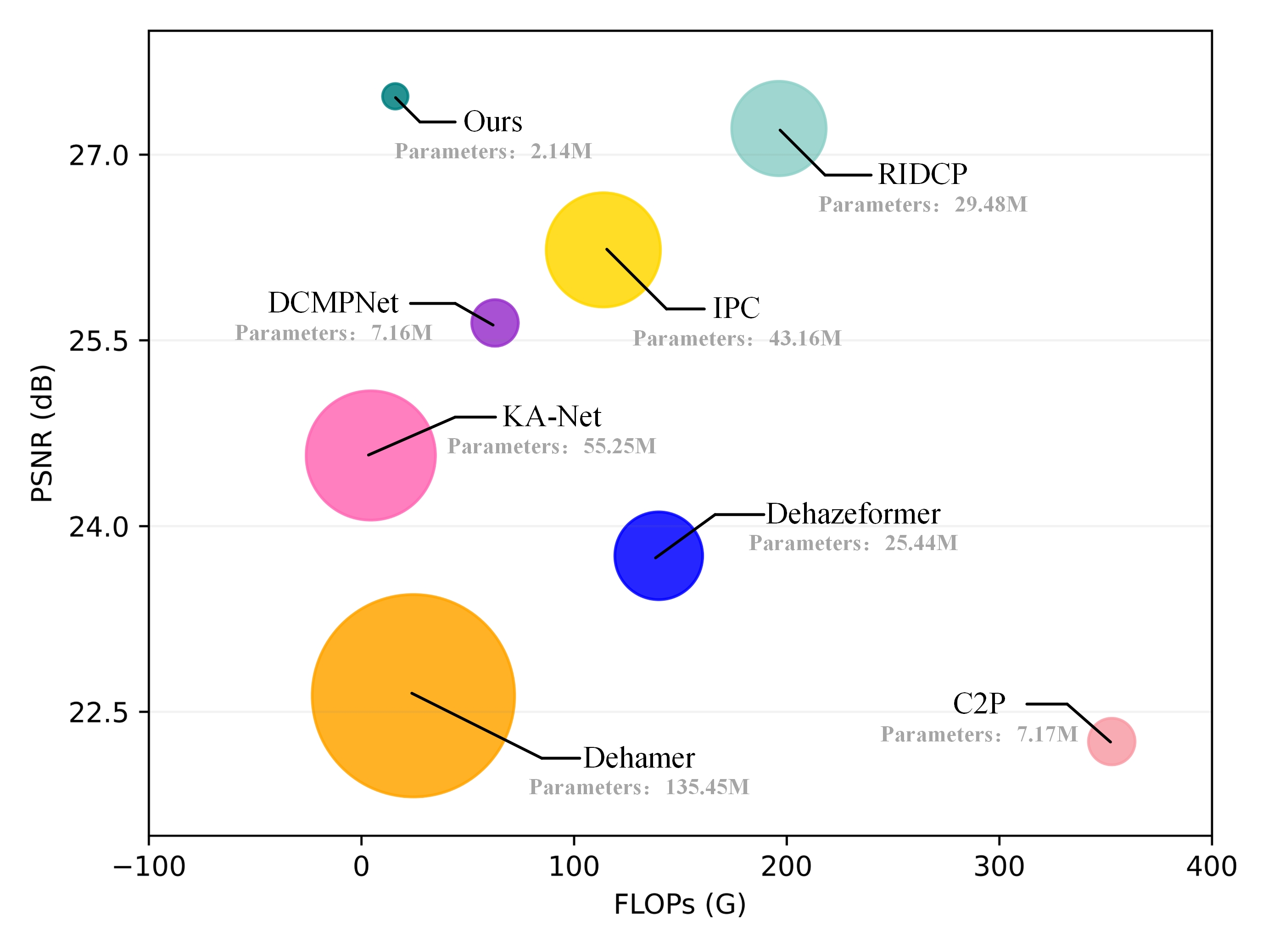}\vspace{-3mm}
	\end{center}
	\caption{Model complexity analysis. The x-axis in the figure represents the FLOPs (in billions) obtained by the model on a 256×256 image input, the y-axis represents PSNR(dB), and the bubble size indicates the number of parameters.}\vspace{-2mm}
	\label{fig7}
\end{figure}

This section presents a comparison of the parameter count and computational complexity (FLOPs) across different methods. As illustrated in Figure \ref{fig7}, our method achieves the best performance with the lowest parameter count and ranks second in FLOPs, demonstrating its efficiency in both model size and computational cost.


\begin{figure*}[t]
	\begin{center}
		\includegraphics[width=0.95\linewidth]{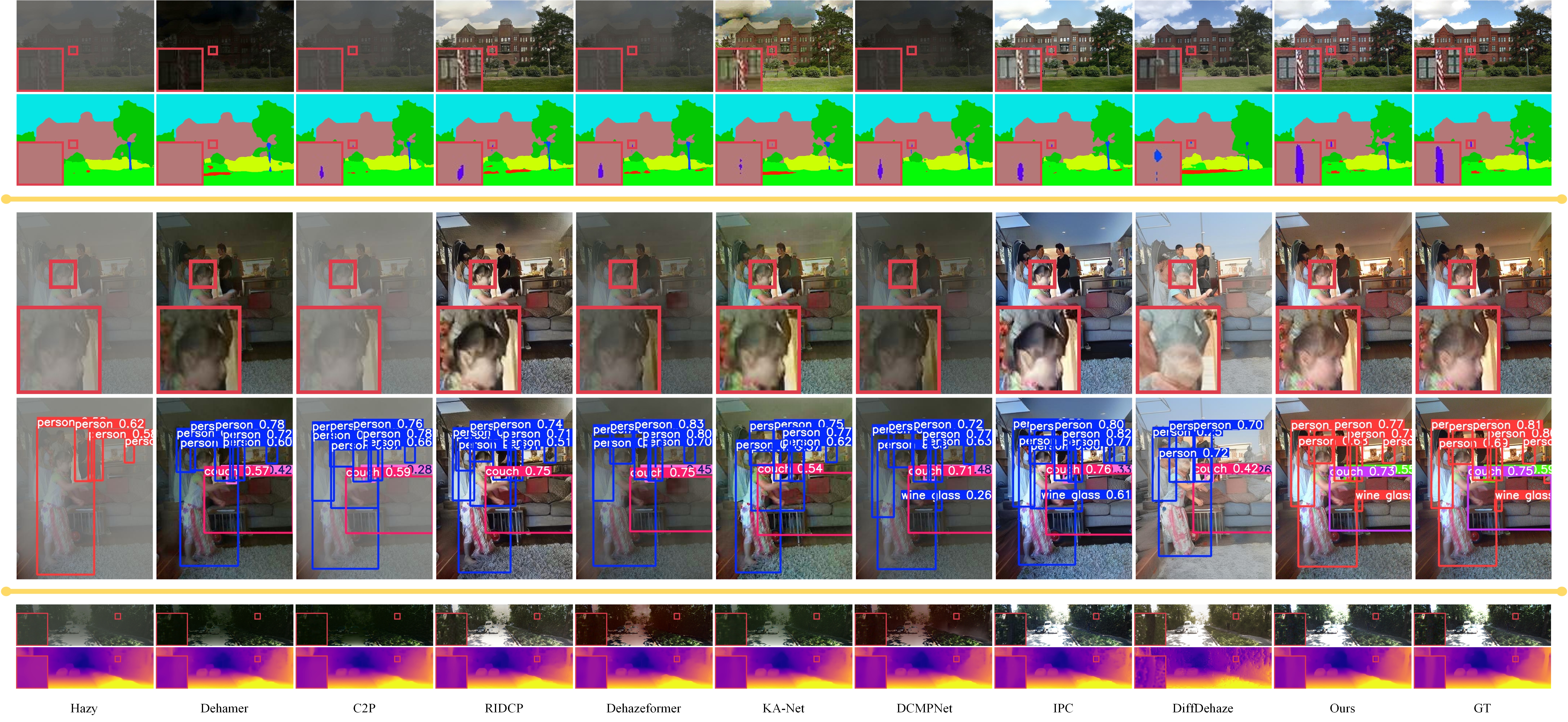}\vspace{-3mm}
	\end{center}
	\caption{Visual comparison with state-of-the-art methods on Setting 1. The results are organized into three parts according to downstream tasks, each comprising four rows: dehazing results (rows 1) and corresponding downstream task outputs (rows 2). Input images are taken from ADE20K, COCO, and KITTI.}\vspace{-2mm}
	\label{fig10}
\end{figure*}
\subsection{2.Comparison with State-of-the-arts}
To validate the effectiveness of the proposed method, we compared it with eight state-of-the-art methods: Dehamer \cite{23}, C2P \cite{16}, RIDCP \cite{37}, Dehazeformer \cite{24}, KA-Net \cite{38}, DCMPNet \cite{17}, IPC \cite{39}, and DiffDehaze \cite{19}. Specifically, we design three experimental settings: \textbf{Setting 1} trains specific downstream tasks using the dehazing results from each comparison method; \textbf{Setting 2} directly inputs the dehazed results into downstream task networks for testing; \textbf{Setting 3} fine-tunes the dehazing network based on the performance feedback from the downstream tasks.


\textbf{Setting 1: }To ensure fairness, for other dehazing methods, we individually optimize each downstream task based on their dehazed results to enable better adaptation. In contrast, our method does not require any retraining or fine-tuning of the downstream tasks. To evaluate whether our method can still achieve significant performance improvements under this constraint, we compare it with those methods that require downstream task retraining. The qualitative and quantitative analyses for Setting 1 have been presented in the main text, with additional visualizations shown in Figure \ref{fig10}, further demonstrating the strong generalization and adaptability of our method to various downstream tasks.

\begin{figure*}[t!]
	\begin{center}
		\includegraphics[width=0.95\linewidth]{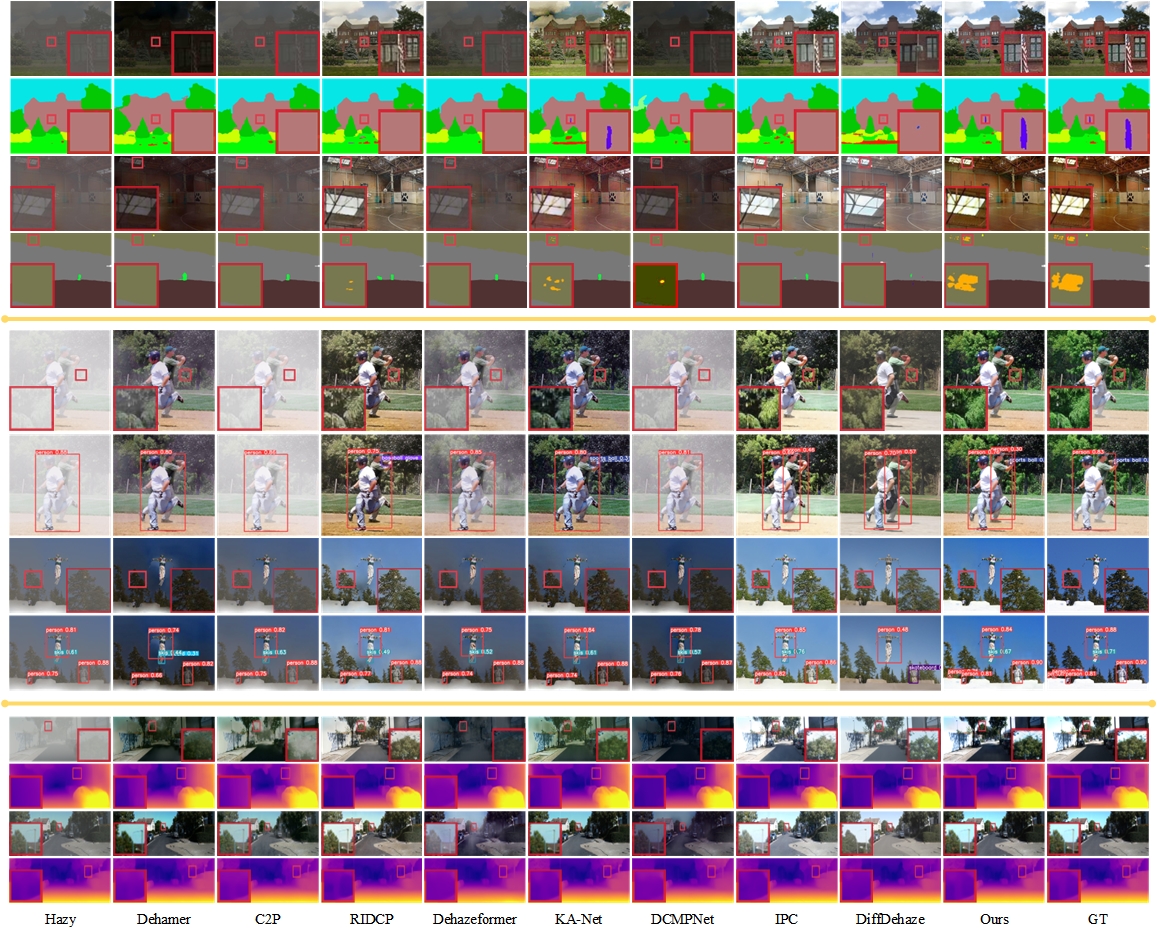}\vspace{-3mm}
	\end{center}
	\caption{Visual comparison with state-of-the-art methods on Setting 2. The results are organized into three parts according to downstream tasks, each comprising four rows: dehazing results (rows 1 and 3) and corresponding downstream task outputs (rows 2 and 4). Input images are taken from ADE20K, COCO, and KITTI.}\vspace{-2mm}
	\label{fig8}
\end{figure*}

\textbf{Setting 2: }We directly evaluate the dehazing results from existing dehazing methods using multiple pretrained downstream task models, without any additional adaptation or fine-tuning. This setting is designed to assess how well dehazing methods generalize to various downstream tasks when no specific optimization is applied to either the dehazing networks or the downstream models.
As shown in Figure \ref{fig8}, the zoomed-in regions reveal that our method better preserves image detail. In the semantic segmentation area, our method performs better at segmenting windows and flagpoles in the image. In the object detection area, compared to other methods, our approach achieves better dehazing effects and higher detection accuracy. In the depth estimation area, our method preserves more detail. We further perform quantitative evaluation using objective metrics. As shown in Table \ref{Table1}, most methods show suboptimal performance in downstream task networks. This indicates that dehazing networks or downstream task networks that are not specifically optimized struggle to meet the requirements of multiple downstream tasks simultaneously. In contrast, our proposed method demonstrates excellent performance across multiple downstream tasks.

\begin{figure*}[t!]
	\begin{center}
		\includegraphics[width=0.95\linewidth]{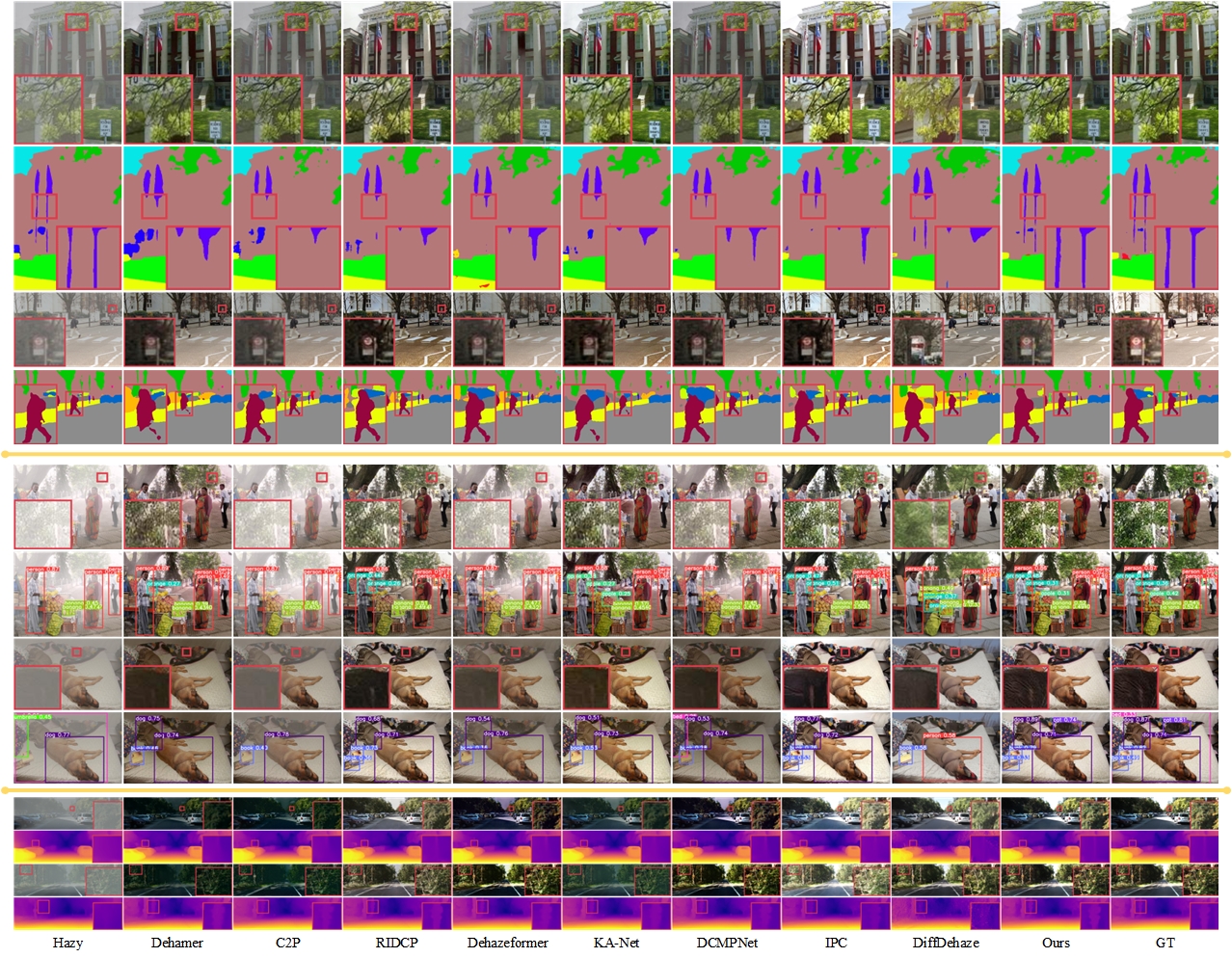}\vspace{-3mm}
	\end{center}
	\caption{Visual comparison with state-of-the-art methods on Setting 3. The results are organized into three parts according to downstream tasks, each comprising four rows: dehazing results (rows 1 and 3) and corresponding downstream task outputs (rows 2 and 4). Input images are taken from ADE20K, COCO, and KITTI.}\vspace{-2mm}
	\label{fig9}
\end{figure*}

\textbf{Setting 3: } In order to adapt the outputs of dehazing networks to the inputs of downstream tasks, we fine-tune the existing dehazing networks based on the performance of downstream tasks. In contrast, our method does not require any retraining or fine-tuning of the dehazing network.This strategy aims to assess whether our dehazing network can achieve better performance solely by relying on two fine-tuning modules, without retraining. To further validate the effectiveness of our method, we compare it with dehazing methods fine-tuned with downstream task loss. As shown in Figure \ref{fig9}, in the semantic segmentation area, our method is able to more accurately segment objects in the image. In the object detection area, our method detects more objects, such as in the 8th row of Figure \ref{fig9}, where our method better identifies the cat in the image. In the depth estimation area, our method retains more detailed information in the depth estimation task. We further conduct quantitative evaluations using objective metrics. As shown in Table \ref{Table2}, despite other dehazing methods actively adapting to the requirements of downstream tasks, our method still demonstrates optimal or near-optimal performance on downstream tasks. In contrast, our method does not require retraining the dehazing model, yet still effectively improves downstream task performance. This further proves that our method can satisfy the requirements of downstream tasks well by simple feature fine-tuning, without requiring the dehazing model to actively adapt to the downstream tasks.

\begin{table*}[ht!]
	\centering\small
	\caption{Comparison of the proposed method across semantic segmentation (SS), object detection (OD), and depth estimation (DE) on Setting 2. Best and runner-up values are highlighted in bold and underlined, respectively.}
	\renewcommand{\arraystretch}{1.1}
	\resizebox{0.95\textwidth}{!}{%
		\begin{tabular}{l|c|cc|cccc|ccc}
			\hline
			\multirow{2.1}{*}{Methods} & \multicolumn{1}{c|}{SS} & \multicolumn{2}{c|}{OD} & \multicolumn{4}{c|}{DE (Error Metric ↓)} & \multicolumn{3}{c}{DE (Accuracy Metric ↑)} \\
			\cline{2-11}
			& mIoU↑ & mAP↑ & mAP50-95↑ & AbsRel & SqRel & RMSE & RMSElog& $\delta{<}1.25 $ & $\delta{<}1.25^2 $ & $\delta{<}1.25^3 $ \\
			\hline
			Dehamer \cite{23} & 43.46 & 43.8 & 25.6 & 0.123 & 0.894 & 5.102 & 0.202 & 0.853 & 0.953 & 0.980 \\
			C2P \cite{16} & 44.10 & 45.2 & 29.2 & 0.152 & 1.143 & 5.782 & 0.235 & 0.792 & 0.925 & 0.966 \\
			RIDCP \cite{37} & 43.73 & 47.5 & 30.4 & 0.104 & 0.716 & 4.414 & 0.179 & 0.890 & \underline{0.966} & \underline{0.984} \\
			Dehazeformer \cite{24} & 44.69 & 46.8 & 30.0 & 0.125 & 0.850 & 5.027 & 0.202 & 0.849 & 0.963 & 0.980 \\
			KA-Net \cite{38} & 43.67 & 44.3 & 26.6 & 0.117 & 0.826 & 4.776 & 0.193 & 0.867 & 0.957 & 0.982 \\
			DCMPNet \cite{17} & 45.08 & 46.7 & 29.9 & 0.116 & 0.799 & 4.912 & 0.195 & 0.862 & 0.956 & 0.981 \\
			DiffDehaze \cite{19} & 39.76 & 38.1 & 23.7 & 0.112 & 0.834 & 4.594 & 0.185 & 0.881 & 0.963 & 0.983 \\
			IPC \cite{39} & \underline{45.35} & \underline{49.3} & \underline{31.6} & \underline{0.101} & \underline{0.690} & \underline{4.315} & \underline{0.175} & \underline{0.897} & \textbf{0.967} & \textbf{0.985} \\
			\textbf{Ours} & \textbf{50.34} & \textbf{54.7} & \textbf{35.7} & \textbf{0.099} & \textbf{0.662} & \textbf{4.310} & \textbf{0.174} & \textbf{0.898} & \textbf{0.967} & \underline{0.984} \\
			\hline
	\end{tabular}}
	\label{Table1}
\end{table*}

\begin{table*}[ht!]
	\centering\small
	\caption{Comparison of the proposed method across semantic segmentation (SS), object detection (OD), and depth estimation (DE) on Setting 3. Best and runner-up values are highlighted in bold and underlined, respectively.}
	\renewcommand{\arraystretch}{1.1}
	\resizebox{0.95\textwidth}{!}{ 
		\begin{tabular}{l|c|cc|cccc|ccc}
			\hline
			\multirow{2.1}{*}{Methods} & \multicolumn{1}{c|}{SS} & \multicolumn{2}{c|}{OD} & \multicolumn{4}{c|}{DE (Error Metric ↓)} & \multicolumn{3}{c}{DE (Accuracy Metric ↑)} \\
			\cline{2-11}
			& mIoU↑ & mAP↑ & mAP50-95↑ & AbsRel & SqRel & RMSE & RMSElog& $\delta{<}1.25 $ & $\delta{<}1.25^2 $ & $\delta{<}1.25^3 $ \\
			\hline
			Dehamer \cite{23} & 44.32 & 49.5 & 31.8 & 0.109 & 0.852 & 4.762 & 0.192 & 0.882 & 0.958 & 0.981 \\
			C2P \cite{16} & 47.14 & 52.2 & 33.4 & 0.119 & 0.884 & 4.892 & 0.196 & 0.862 & 0.951 & 0.980 \\
			RIDCP \cite{37} & \underline{49.18} & 51.6 & 33.1 & 0.101 & 0.690 & 4.320 & \underline{0.175} & 0.897 & \underline{0.966} & \textbf{0.985} \\
			Dehazeformer \cite{24} & 47.41 & 50.1 & 32.2 & 0.110 & 0.756 & 4.554 & 0.185 & 0.878 & 0.962 & 0.983 \\
			KA-Net \cite{38} & 45.28 & 51.2 & 31.9 & 0.107 & 0.750 & 4.467 & 0.189 & 0.881 & 0.958 & 0.982 \\
			DCMPNet \cite{17} & 47.32 & \underline{53.5} & \underline{34.9} & 0.106 & 0.713 & 4.453 & 0.186 & 0.876 & 0.960 & 0.982 \\
			DiffDehaze \cite{19} & 42.73 & 41.2 & 25.9 & 0.118 & 0.863 & 4.991 & 0.201 & 0.891 & 0.964 & 0.983 \\
			IPC \cite{39} & 44.60 & 52.0 & 33.6 & \underline{0.100} & \underline{0.689} & \textbf{4.292} & \textbf{0.174} & \textbf{0.901} & 0.965 & 0.983 \\
			\textbf{Ours} & \textbf{50.34} & \textbf{54.7} & \textbf{35.7} & \textbf{0.099} & \textbf{0.662} & \underline{4.310} & \textbf{0.174} & \underline{0.898} & \textbf{0.967} & \underline{0.984} \\
			\hline
	\end{tabular}}
	\label{Table2}
\end{table*}

\begin{table*}[t]
	\centering\small
	\caption{Comparison of the proposed method across semantic segmentation (SS), object detection (OD), and depth estimation (DE). Bold indicates the best result.}
	\renewcommand{\arraystretch}{1.1}
	\resizebox{0.95\textwidth}{!}{%
		\begin{tabular}{l|c|cc|cccc|ccc}
			\hline
			\multirow{2.1}{*}{$\lambda$} & \multicolumn{1}{c|}{SS} & \multicolumn{2}{c|}{OD} & \multicolumn{4}{c|}{DE (Error Metric ↓)} & \multicolumn{3}{c}{DE (Accuracy Metric ↑)} \\
			\cline{2-11}
			& mIoU↑ & mAP↑ & mAP50-95↑ & AbsRel & SqRel & RMSE & RMSElog& $\delta{<}1.25 $ & $\delta{<}1.25^2 $ & $\delta{<}1.25^3 $ \\
			\hline
			0 & 50.21 & 53.4 & 34.8 & 0.101 & 0.691 & 4.44 & 0.184 & 0.879 & 0.956 & 0.976 \\
			0.01 & 50.29 & 53.9 & 35.1 & 0.101 & 0.683 & 4.41 & 0.179 & 0.886 & 0.963 & 0.981 \\
			0.1 & \textbf{50.34} & \textbf{54.7} &\textbf{35.7} & \textbf{0.099} & \textbf{0.662} & \textbf{4.31} & \textbf{0.174} & \textbf{0.898} & \textbf{0.967} & \textbf{0.984} \\
			1 & 50.33 & 54.4 & 35.5 & \textbf{0.099} & 0.683 & 4.33 & 0.175 & 0.897 & \textbf{0.967} & \textbf{0.984} \\
			\hline
		\end{tabular}
	}
	\label{Table3}
\end{table*}
\begin{table*}[t]
	\centering\small
	\caption{Comparison of the proposed method across semantic segmentation (SS), object detection (OD), and depth estimation (DE). Bold indicates the best result.}
	\renewcommand{\arraystretch}{1.1}
	\resizebox{0.95\textwidth}{!}{%
		\begin{tabular}{l|c|cc|cccc|ccc}
			\hline
			\multirow{2.1}{*}{$\gamma$} & \multicolumn{1}{c|}{SS} & \multicolumn{2}{c|}{OD} & \multicolumn{4}{c|}{DE (Error Metric ↓)} & \multicolumn{3}{c}{DE (Accuracy Metric ↑)} \\
			\cline{2-11}
			& mIoU↑ & mAP↑ & mAP50-95↑ & AbsRel & SqRel & RMSE & RMSElog& $\delta{<}1.25 $ & $\delta{<}1.25^2 $ & $\delta{<}1.25^3 $ \\
			\hline
			0 & 50.18 & 53.5 & 34.9 & 0.102 & 0.691 & 4.43 & 0.183 & 0.878 & 0.955 & 0.976 \\
			0.01 & \textbf{50.34} & \textbf{54.7} & \textbf{35.7} & \textbf{0.099} & \textbf{0.662} & \textbf{4.31} & \textbf{0.174} & \textbf{0.898} & \textbf{0.967} & \textbf{0.984} \\
			0.1 & 50.31 & 54.5 & 35.6 & 0.101 & 0.664 & 4.33 & 0.177 & 0.896 & 0.964 & \textbf{0.984} \\
			1 & 50.32 & \textbf{54.7} & 35.6 & \textbf{0.099} & 0.663 & 4.32 & \textbf{0.174} & 0.897 & 0.965 & \textbf{0.984} \\
			\hline
		\end{tabular}
	}
	\label{Table4}
\end{table*}

\subsection{3.Parameter Selection and Analysis}

In the main body of this paper, we set the hyperparameter $\lambda$ in Eq. (1) and the hyperparameter $\gamma$ in Eq. (2) to 0.1 and 0.01, respectively. Here, we fix one hyperparameter and vary the other to verify the effectiveness of $\lambda = 0.1$ and $\gamma = 0.01$. Tables \ref{Table3} and  \ref{Table4} show the changes in the metrics brought by the variation of each parameter.

\begin{equation}
	\footnotesize
	\begin{array}{l}
		\ell_{\text{dehaze}} = \|J'_w(x) - J(x)\|_1 + \\
		\quad \lambda \sum_{v=1}^{n} \beta_v \frac{\|VGG_v(J(x)) - VGG_v(J'_w(x))\|_1}{\|VGG_v(J'_w(x)) - VGG_v(\tilde{J}(x))\|_1}
	\end{array}
\end{equation}

\begin{equation}
	\footnotesize
	\ell_{total} = \ell_{dehaze} + \ell_{mcr} + \gamma \ell_{down}
\end{equation}

\textbf{Selection and Analysis of Parameter $\lambda$: }We investigate the impact of different values of $\lambda$ in the range [0,1] on the model's performance. Table \ref{Table3} shows the changes in metrics for various downstream tasks at different values of $\lambda$. From the table, it can be seen that when $\lambda = 0.1$, the model performs best in downstream tasks.

\textbf{Selection and Analysis of Parameter $\gamma$: } We investigate the impact of different values of $\gamma$ in the range [0,1] on the model's performance. Table \ref{Table4} shows the changes in metrics for various downstream tasks at different values of $\gamma$. From the table, it can be seen that when $\gamma = 0.01$, the model performs best in downstream tasks.

\end{document}